# Ontology-based Fuzzy Markup Language Agent for Student and Robot Co-Learning

Chang-Shing Lee, Mei-Hui Wang, Tzong-Xiang Huang
Li-Chung Chen, Yung-Ching Huang, Sheng-Chi Yang
Dept. of Computer Science and Information Engineering
National University of Tainan
Tainan, Tawain
leecs@mail.nutn.edu.tw

Chien-Hsun Tseng, Pi-Hsia Hung
Dept. of Education
National University of Tainan
Tainan, Tawain
tjs1219@mail2000.com.tw

Naoyuki Kubota
Dept. of System Design
Tokyo Metropolitan University
Tokyo, Japan
kubota@tmu.ac.jp

***Abstract*-An intelligent robot agent based on domain ontology, machine learning mechanism, and Fuzzy Markup Language (FML) for students and robot co-learning is presented in this paper. The machine-human co-learning model is established to help various students learn the mathematical concepts based on their learning ability and performance. Meanwhile, the robot acts as a teacher's assistant to co-learn with children in the class. The FML-based knowledge base and rule base are embedded in the robot so that the teachers can get feedback from the robot on whether students make progress or not. Next, we inferred students' learning performance based on learning content's difficulty and students' ability, concentration level, as well as teamwork sprit in the class. Experimental results show that learning with the robot is helpful for disadvantaged and below-basic children. Moreover, the accuracy of the intelligent FML-based agent for student learning is increased after machine learning mechanism.**

## I. INTRODUCTION

Ontology model can provide knowledge representation and reasoning capabilities for machines to solve a task as well as to allow semantic interoperability between systems or agents [1]. Owning to the rapid advance in artificial intelligence (AI), Sophia, a social humanoid robot, came to the world in 2015. She was programmed to give pre-written responses to specific questions and also became the first ever to be granted a full Saudi Arabian citizenship in 2017 [2]. Additionally, Liu et al. [3] proposed a fuzzy ontology representation model to express common fuzzy knowledge. Meditskos and Kompatsiaris [4] presented iKnow to capitalize on the use of OWL ontological knowledge to capture domain relationships between low-level observations and high-level activities. Lee et al. used the ontology to represent the knowledge of patent technology requirement evaluation and recommendation [5] as well as proposed a type-2 fuzzy ontology for personal diabetic-diet recommendation [6].

Fuzzy Markup Language (FML) is a specific purpose markup language based on XML to describe the structure and behavior of a fuzzy system independently of the hardware architecture [13-14]. Since May 2016, FML has become one of the IEEE standards [15-16] and has been applied to a lot of researches like game of Go [11, 17], diet assessment [14, 18, 22], and so on. Nowadays, human and machine co-learning is an important topic for current societies. One way to provide a robot with such learning capability is to use machine learning [8]. Learning explores and understands the learning process of humans, and machine learning studies how algorithms learn from data [9]. Jain et al. [10] proposed an artificial-based student learning evaluation tool to test with students from undergraduate courses. There exists some FML-based real-world applications to persons' learning. For example, Lee et al. proposed a FML-based intelligent adaptive assessment platform for learning materials recommendation [20], and an online self-learning platform construction based on genetic FML (GFML) and item response theory (IRT) agent [21]. They also proposed a FML-based dynamic assessment agent for human-machine cooperative system on game of Go [11]. By combining particle swarm optimization (PSO) with FML, called PFML, Lee et al. applied to human and machine co-learning on game of Go [12] and student learning performance evaluation [19].

Advancement in technology is bringing robots into interpersonal aspects of student's learning [7]; therefore, including the robot into the class to co-learn with humans has been a trend for recent years. This paper brings the robots Palro, developed by FUJISOFT Japan and Zenbo, developed by ASUS, Taiwan, into an elementary school to co-learn mathematics with four-grade children. The objective of this paper is to represent the knowledge of the robot for student and robot co-learning. We first construct the student learning performance ontology for the robot agent to predict their learning performance. Then, we construct the student and robot co-learning ontology to recommend students for suitable learning contents. After that, we use FML to describe the knowledge base and rule base for the constructed ontologies. Finally, we apply the developed robot agent to the four-grade students for learning mathematical concepts of number line and groups of numbers. We also use machine learning techniques to optimize the involved student learning performance using GFML [17-18] and PFML [19]. The experiments show that the proposed ontology-based fuzzy markup language agent is feasible for student and robot co-learning.

The remainder of this paper is organized as follows: Section II introduces the ontology model for student and robot co-learning. Section III describes the fuzzy markup language agent, including the proposed system structure as well as knowledge base, rule base, and optimization model for the proposed student and robot co-learning. The experimental results are shown in Section IV and conclusions are given in Section V.

The authors would like to thank the financially support sponsored by the Ministry of Science and Technology of Taiwan under the grants MOST 106-3114-E-024-001 and 106-2221-E-024-019.

## II. ONTOLOGY MODEL FOR STUDENT AND ROBOT CO-LEARNING

### A. *Student Learning Performance Ontology for Robot Agent*

The structure of student learning performance ontology based on Fuzzy Markup Language (FML) for robot agent reasoning is shown in Fig.1. The domain name is defined as FML output variable which connects to the FML linguistic concepts of the output fuzzy variable. There are some FML input variables, for example, FML input variable 1, FML input variable 2, FML input variable 3, …, and FML input variable *M*, to connect to the linguistic concepts of FML input variables. Each FML input variable contains some linguistic concepts. The linguistic concept of FML input variable contains some attributes, such as *Area*, *Grade*, *Subject*, and so on. In addition, some operations are also defined in the linguistic concept, for example, operation *Recommend*.

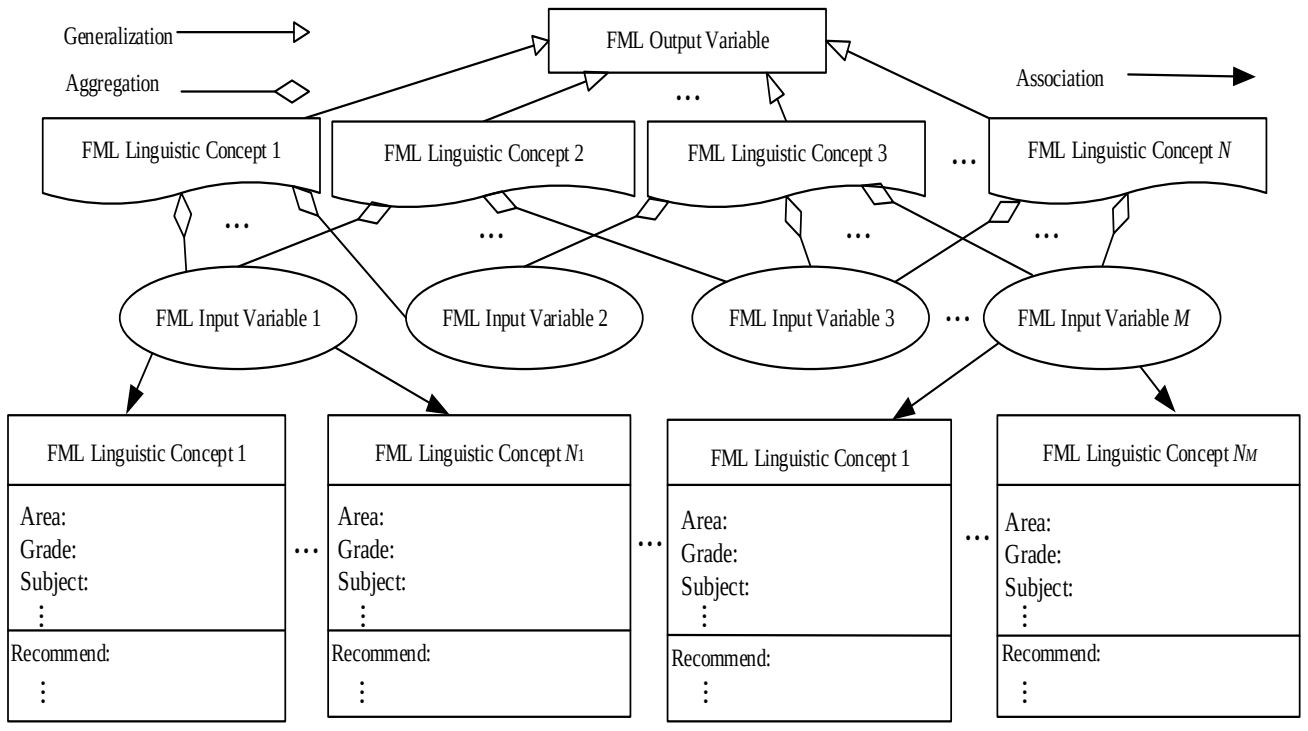

Fig. 1. Structure of FML-based student learning performance ontology.

### B. *An Instance of Student Learning Performance Ontology*

Fig. 2 shows an instance of FML-based student learning performance ontology. The domain name for FML output variable is *Student Learning Performance* (*SLP*), there are four FML input variables in the ontology, including: *Student Ability* (*SA*), *Learning Content Difficult* (*LCD*), *Student Concentration Level* (*SCL*), and *Student Teamwork Spirit* (*STS*). There are five linguistic concepts defined in FML output variable, including *FallBehind*, *Insufficient*, *Basic*, *Good*, and *Excellent*. Each FML input variable contains four linguistic concepts in this paper. For instance, *LCD* has an association relations with *VeryEasy*, *Easy*, *Average*, and *Hard*. Each linguistic concept contains some attributes like *Area*, *Grade*, and *Subject*, as well as operations like *Recommend*.

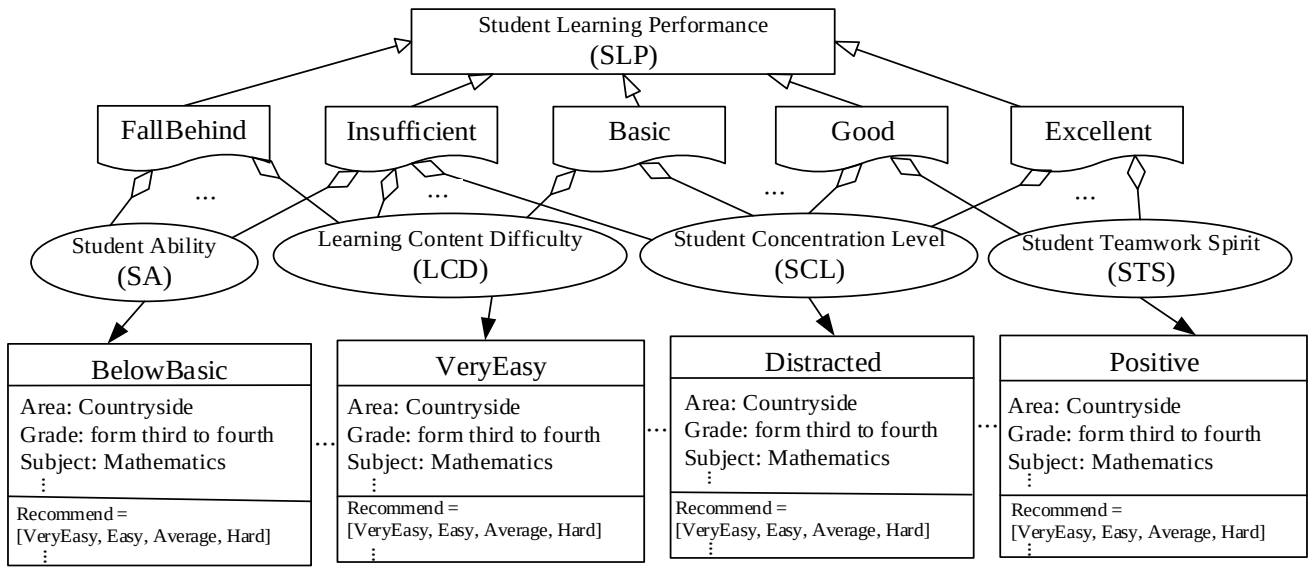

Fig. 2. An instance of FML-based student learning performance ontology.

### C. *Student and Robot Co-Learning Ontology*

Fig. 3 shows student and robot co-learning ontology. The developed robot agent predicts the involved students' learning performance in the class according to the constructed student learning performance ontology shown in Fig. 2. Next, according to the feedback of the students' learning performance and students' ability, the robot agent provides students with suitable learning contents for their next study. Fig. 3 also shows partial learning content ontology about the concepts of number line and groups of numbers. Categories *Number and Calculation* and *Quantity and Measurement* have concepts *Number Line*, *Distance*, *Four fundamental operations of arithmetic*, …, etc. For example, before knowing the concepts of *Number Line*, students should know the concepts of *Positive Number* which includes the concept of *Positive Integer*.

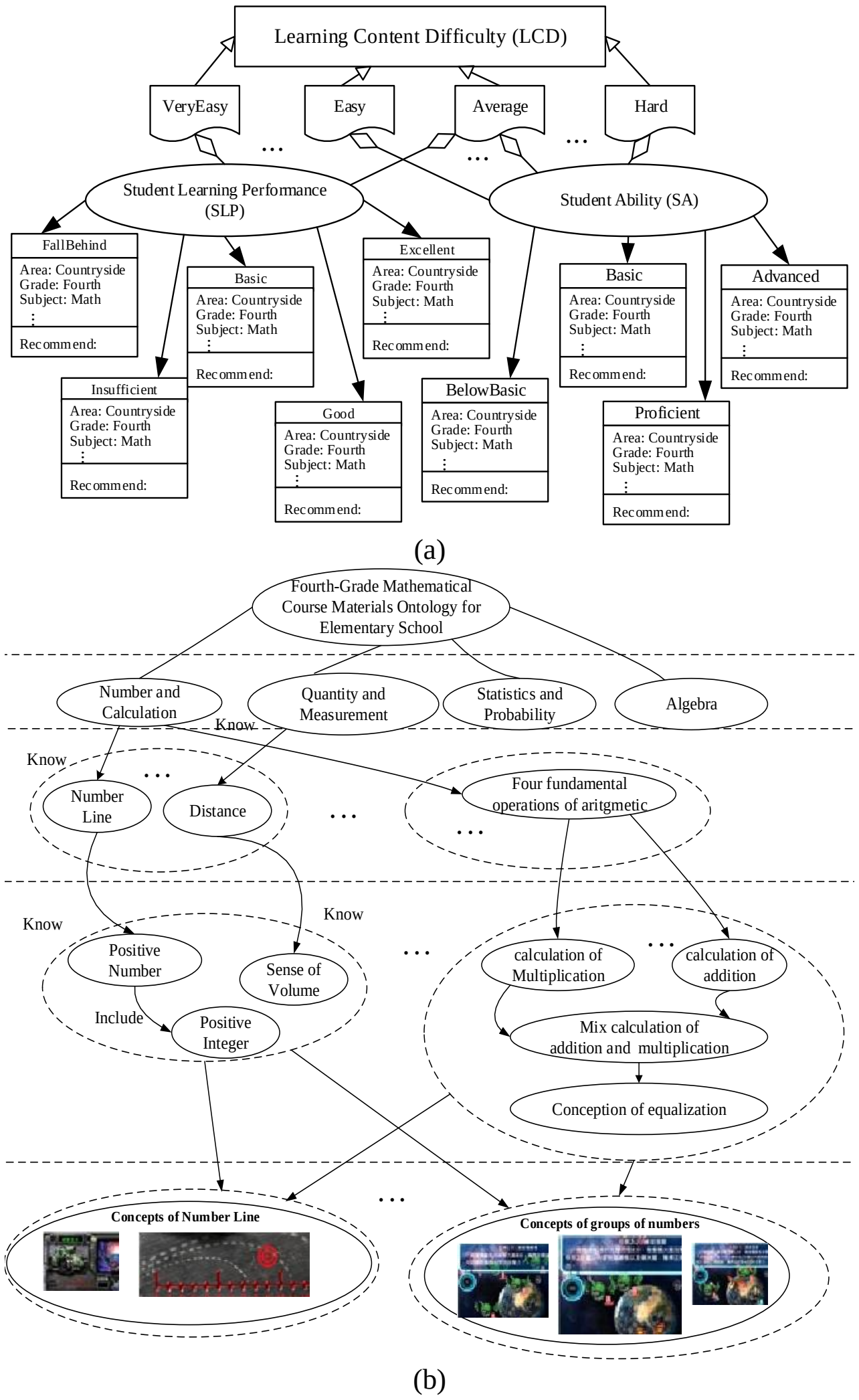

Fig. 3. (a) Student and robot co-learning ontology and (b) an example of learning content ontology of the fourth-grade number line and groups of numbers.

## III. FML-BASED INTELLIGENT AGENTS FOR STUDENT AND ROBOT CO-LEARNING

### *A. System Structure for Student and Robot Co-Learning*

We propose the FML-based intelligent agents, including an *ontology agent*, a *teaching assistant agent*, a *co-learning learning*, an *assessment agent*, and a *recommendation agent*, for student and robot co-learning in this section. Fig. 4 shows the structure of FML-based intelligent robot agents for student learning performance assessment and learning content recommendation. The publisher finds domain experts to write the textbook for teacher's teaching and student's learning after the government, for example, Ministry of Education, defines the learning outline for different grades of students. The ontology agent constructs the domain ontology based on the textbook for the teaching assistant agent. In addition, the co-learning agent helps teacher teach students in the class and the assessment agent classifies the student learning performance into five categories, including *FallBehind*, *Insufficient*, *Basic*, *Good*, and *Excellent*. Finally, the recommendation agent helps teachers and students choose suitable learning contents for their further study and learning. Fig. 5 shows the communication structure among the developed learning contents, students, and the robot Palro which communicates with the server via the developed robot socket client.

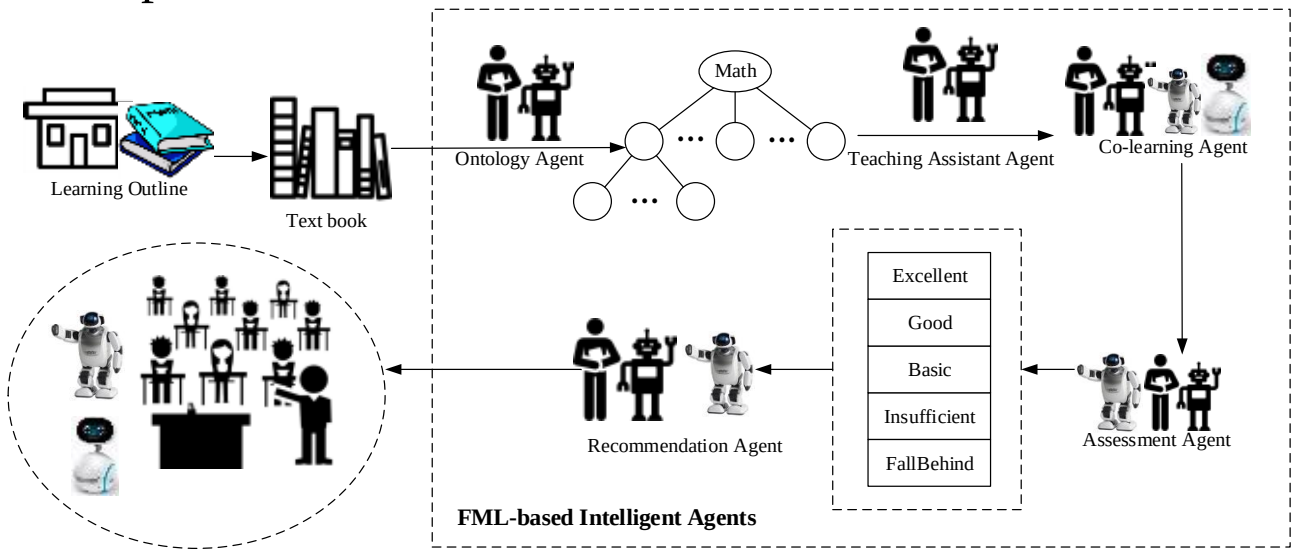

Fig. 4. Structure of FML-based intelligent robot agent for student learning performance assessment and learning content recommendation.

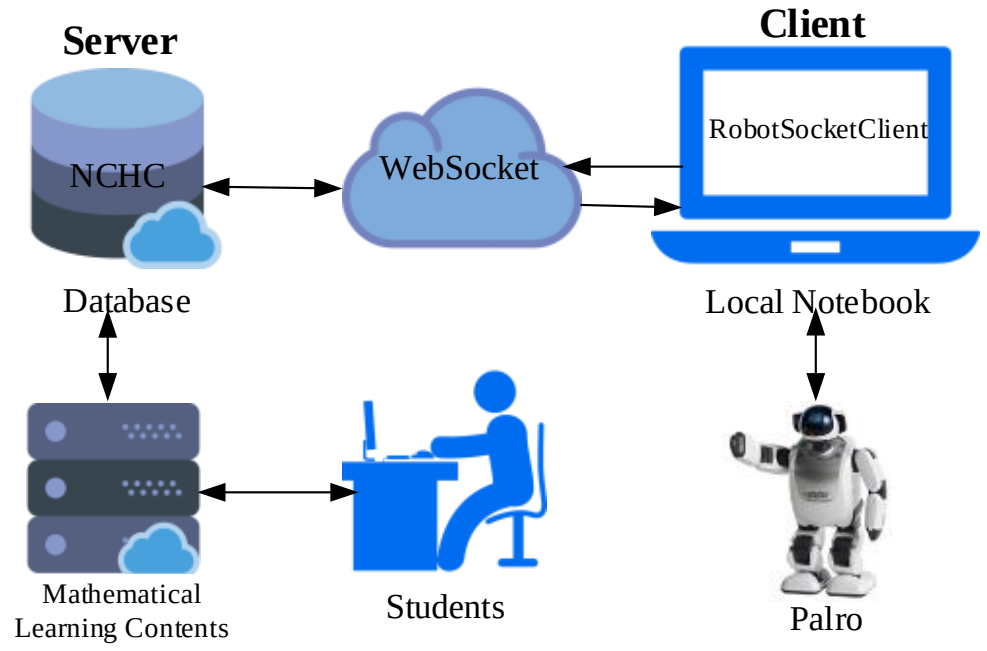

Fig. 5. Communication structure among the developed learning contents, students, and Palro.

### *B. Knowledge Base for Human and Robot Co-Learning*

In this paper, we propose a FML Robot Agent for *Student Learning Performance Prediction*, the knowledge base for FML input variables are defined as follows: (1) *Student Ability* (*SA*) = {***BelowBasic***, ***Basic***, ***Proficient***, ***Advanced***}; (2) *Learning Content Difficulty* (*LCD*) = {***VeryEasy***, ***Easy***, ***Average***, ***Hard***}; (3) *Student Concentration Level* (*SCL*) = {***Distracted***, ***Nonfocused***, ***Focused***, ***Absorbed***}; (4) *Student Teamwork Spirit* (STS) = {***Passive***, ***Normal***, ***Initiative***, ***Positive***}. In addition, we define the knowledge base for FML output variable *Student Learning Performance* (*SLP*) = {***FallBehind***, ***Insufficient***, ***Basic***, ***Good***, ***Excellent***}. Figs. 6(a)-6(e) show the fuzzy sets for fuzzy variables *SA*, *LCD*, *SCL*, *STS*, and *SLP*, respectively.

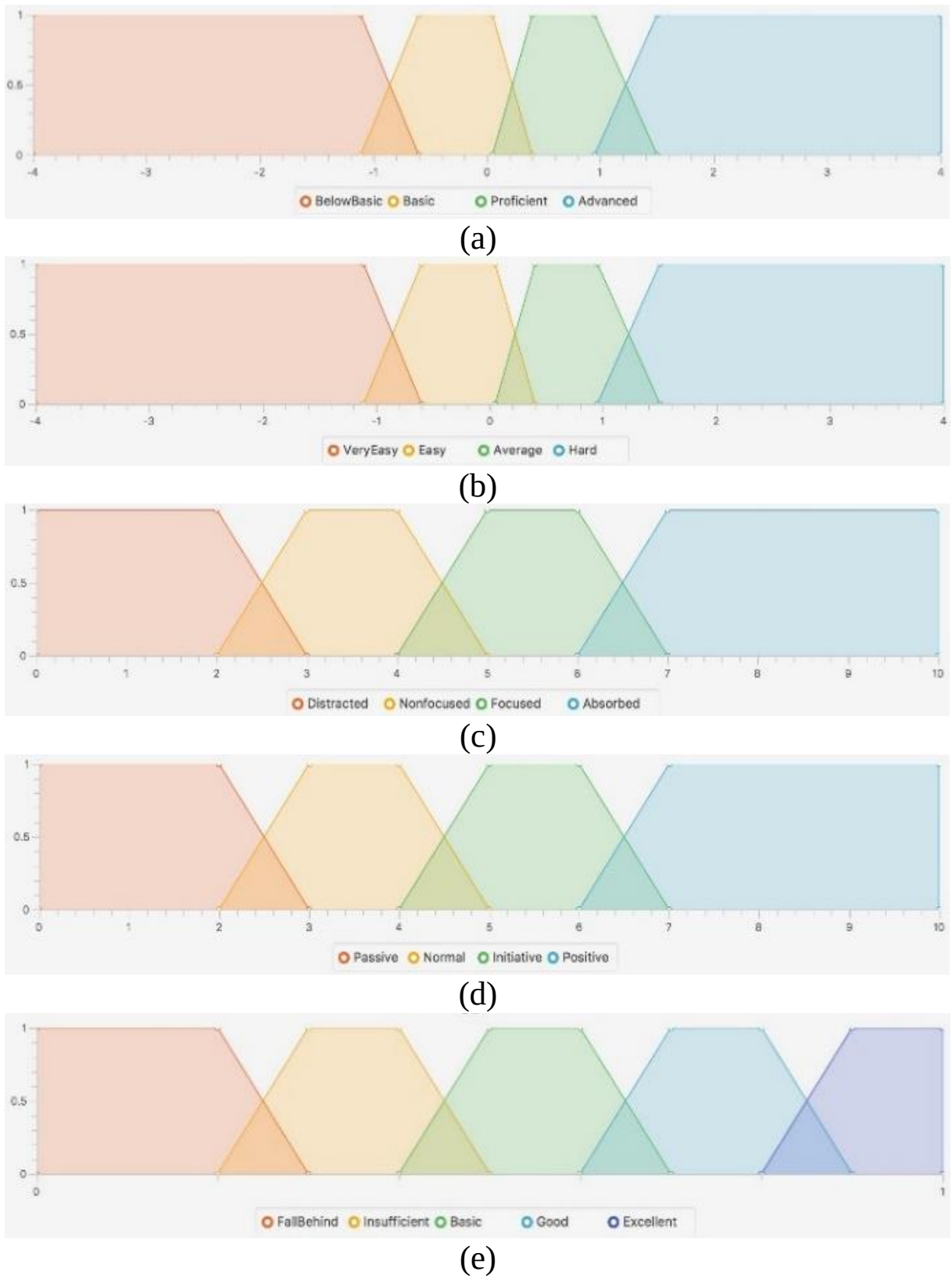

Fig. 6. FML input variables (a) *SA*, (b) *LCD*, (c) *SCL*, (d) *STS*, and (e) FML output variable *SLP*.

### *C. Rule Base for Human and Robot Co-Learning*

The FML *robot agent* first predicts students' learning performance based on the knowledge base, and then provides suitable learning contents to students for next study according to the students' learning performance. In 2017, we developed the mathematical learning contents using PHP and Java languages to construct the learning-content server which allows students to surf on it to learn in the class. Table I shows partial fuzzy rules and Table II shows partial knowledge base and rule base of FML that predicts students' learning performance.

TABLE I. PARTIAL FUZZY RULES

| No | SA | LCD | SCL | STS | SLP |
|---|---|---|---|---|---|
| 1 | BelowBasic | VeryEasy | Distracted | Passive | FallBehind |
| 2 | BelowBasic | VeryEasy | Distracted | Normal | FallBehind |
| 3 | BelowBasic | VeryEasy | Distracted | Initiative | FallBehind |
| 4 | BelowBasic | VeryEasy | Distracted | Positive | FallBehind |
| 5 | BelowBasic | VeryEasy | Nonfocused | Passive | FallBehind |
| 6 | BelowBasic | VeryEasy | Nonfocused | Normal | FallBehind |
| 7 | BelowBasic | VeryEasy | Nonfocused | Initiative | FallBehind |
| 8 | BelowBasic | VeryEasy | Nonfocused | Positive | Insufficient |
| 9 | BelowBasic | VeryEasy | Focused | Passive | FallBehind |
| 10 | BelowBasic | VeryEasy | Focused | Normal | FallBehind |
| | | | ⋮ | | |
| 250 | Advanced | Hard | Focused | Normal | Excellent |
| 251 | Advanced | Hard | Focused | Initiative | Excellent |
| 252 | Advanced | Hard | Focused | Positive | Excellent |
| 253 | Advanced | Hard | Absorbed | Passive | Excellent |
| 254 | Advanced | Hard | Absorbed | Normal | Excellent |
| 255 | Advanced | Hard | Absorbed | Initiative | Excellent |
| 256 | Advanced | Hard | Absorbed | Positive | Excellent |

TABLE II. PARTIAL KNOWLEDGE BASE AND RULE BASE OF FML

```
<?xml version="1.0" encoding="UTF-8"?>
<fuzzySystem xmlns="http://www.ieee1855.org" name="SLPSystemRB" networkAddress="127.0.0.1">
    <knowledgeBase networkAddress="127.0.0.1">
        <fuzzyVariable name="SA" domainleft="-4" domainright="4" type="Input" accumulation="MAX" defuzzifier="COG" defaultValue="0.0"
        networkAddress="127.0.0.1">
            <fuzzyTerm name="BelowBasic" complement="false">
                <trapezoidShape param1="-4" param2="-4" param3="-1.11" param4="-0.6"/>
            </fuzzyTerm>
            <fuzzyTerm name="Basic" complement="false">
                <trapezoidShape param1="-1.11" param2="-0.6" param3="0.05" param4="0.4"/>
            </fuzzyTerm>
            <fuzzyTerm name="Proficient" complement="false">
                <trapezoidShape param1="0.05" param2="0.4" param3="0.95" param4="1.5"/>
            </fuzzyTerm>
            <fuzzyTerm name="Advanced" complement="false">
                <trapezoidShape param1="0.95" param2="1.5" param3="4" param4="4"/>
            </fuzzyTerm>
        </fuzzyVariable>
        <fuzzyVariable name="LCD" domainleft="-4" domainright="4" type="Input" accumulation="MAX" defuzzifier="COG" defaultValue="0.0"
        networkAddress="127.0.0.1">
            <fuzzyTerm name="VeryEasy" complement="false">
                <trapezoidShape param1="-4" param2="-4" param3="-1.11" param4="-0.6"/>
            </fuzzyTerm>
            <fuzzyTerm name="Easy" complement="false">
                <trapezoidShape param1="-1.11" param2="-0.6" param3="0.05" param4="0.4"/>
            </fuzzyTerm>
            <fuzzyTerm name="Average" complement="false">
                <trapezoidShape param1="0.05" param2="0.4" param3="0.95" param4="1.5"/>
            </fuzzyTerm>
            <fuzzyTerm name="Hard" complement="false">
                <trapezoidShape param1="0.95" param2="1.5" param3="4" param4="4"/>
            </fuzzyTerm>
        </fuzzyVariable>
                                    ⋮
    </knowledgeBase>
                                    ⋮
    <mamdaniRuleBase name="SLPSystemRB" activationMethod="MIN" andMethod="MIN" orMethod="MAX"
    networkAddress="127.0.0.1">
        <rule name="rule-1" andMethod="MIN" orMethod="MAX" connector="AND" weight="1.0"
        networkAddress="127.0.0.1">
                                    ⋮
    </mamdaniRuleBase>
</fuzzySystem>
```

*D. Machine Learning for knowledge base optimalization*

This subsection describes the machine learning methods, including GFML [17, 18, 22] and PFML [19] to optimize the knowledge base of Fuzzy Markup Language. The former is to combine genetic algorithm with FML and the latter is to combine PSO with FML. Fig. 7 shows the FML-based intelligent agents with a machine learning mechanism for students learning. For GFML, three types of genes are defined, including knowledge-based genes (composed of the FML variables' names and the objects with the linguistic terms of their own fuzzy variables), rule-based genes (a collection of the weight of the fuzzy rules), and fuzzy-hedged genes (linguistic terms' hedge of the fuzzy variables) [22]. In this paper, the encoded chromosome has 266 genes, including the knowledge-based genes ($G_1$ to $G_5$), the rule-based genes ($G_6$ to $G_{261}$), and the fuzzy-hedged genes ($G_{262}$ to $G_{266}$).

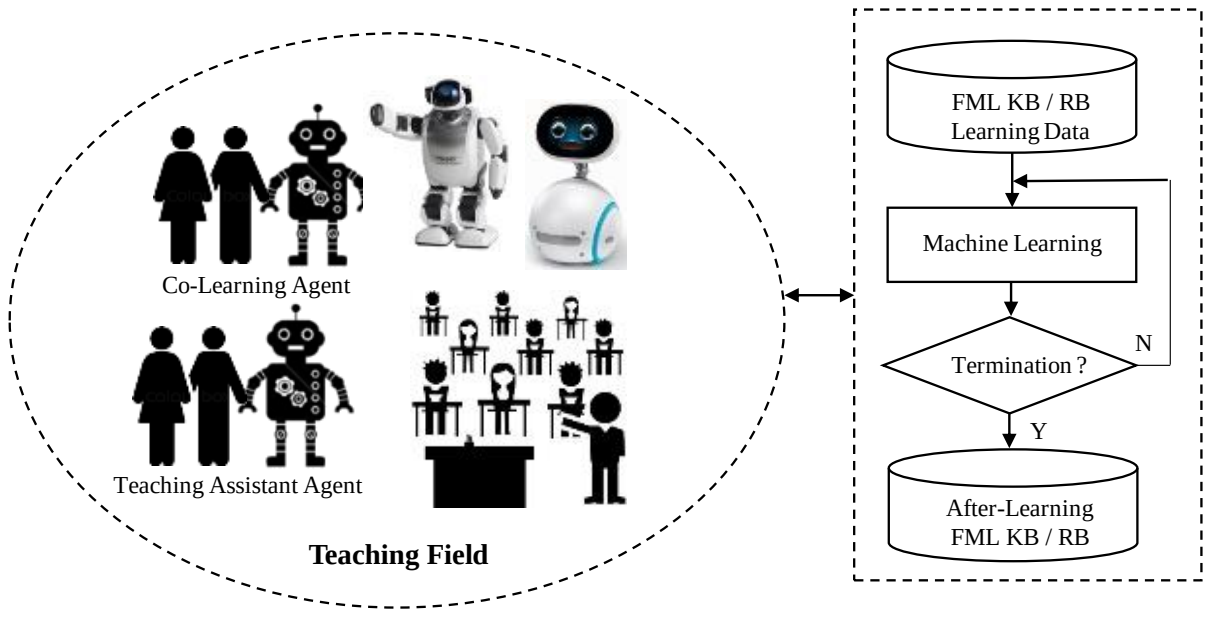

Fig. 7. FML-based intelligent agents with a machine learing mechanism for students learning.

For PFML, the total number of parameters for each particle is 84 in this paper. The parameters of four FML input variables and one FML output variable represent the position of the particle in 5-dimensional space where are optimized by adjusting the moving velocity in order to reach convergence. The domain of the particle in each dimension is bounded in [*domain left*, *domain right*] of the FML variable [19]. In this paper, the domains of from the first to the fifth dimension are [−4, 4], [−4, 4], [0, 10], [0, 10], and [0, 1] to optimize the parameters of FML variables *SA*, *LCD*, *SCL*, *STS*, and *SLP*, respectively.

## IV. EXPERMENTAL RESULTS

There are three parts in the experimental results: (1) Parts 1 and 2 are to test the behavior for FML-based intelligent agent, including an assessment agent and a recommendation agent for student learning performance assessment and learning content recommendation, respectively. (2) Part 3 is to deploy the FML-based intelligent agent with different robots, Palro and Zenbo, to an elementary school for four-grade students that co-learned with the robots about mathematical concepts of number line and types of numbers in Nov. and Dec. 2017, respectively.

*A. Part 1: Student Learning Performance Assessment*

In Part 1 of the experiments, we propose a FML robot agent for *student learning performance assessment*. The knowledge base for FML input variables are defined as follows: (1) *Student Ability* (*SA*) = {***BelowBasic***, ***Basic***, ***Proficient***, ***Advanced***}={**[-4, -4, -1.11, -0.6]**, **[-1.11, -0.6, 0.05, 0.4]**, **[0.05, 0.4, 0.95, 1.5]**, **[0.95, 1.5, 4, 4]**}; (2) *Learning Content Difficulty* (*LCD*) = {***VeryEasy***, ***Easy***, ***Average***, ***Hard*** } ={**[-4, -4, -1.11, -0.6]**, **[-1.11, -0.6, 0.05, 0.4]**, **[0.05, 0.4, 0.95, 1.5]**, **[0.95, 1.5, 4, 4]**}; (3) *Student Concentration Level* (*SCL*) = {***Distracted***, ***Nonfocused***, ***Focused***, ***Absorbed***} ={**[0, 0, 2, 3]**, **[2, 3, 4, 5]**, **[4, 5, 6, 7]**, **[6, 7, 10, 10]**}; (4) *Student Teamwork Spirit* (STS) = {***Passive***, ***Normal***, ***Initiative***, ***Positive***}={**[0, 0, 2, 3]**, **[2, 3, 4, 5]**, **[4, 5, 6, 7]**, **[6, 7, 10, 10]**}. In addition, we define the knowledge base for FML output variable *Student Learning Performance* (*SLP*) = {***FallBehind***, ***Insufficient***, ***Basic***, ***Good***, ***Excellent***}={**[0.0, 0.0, 0.2, 0.3]**, **[0.2, 0.3, 0.4, 0.5]**, **[0.4, 0.5, 0.6, 0.7]**, **[0.6, 0.7, 0.8, 0.9]**, **[0.8, 0.9, 1, 1]**}.

We first simulate 400 records and then use K-fold cross validation method to evaluate the performance. The fitness function is mean square error (MSE). In this paper, $K = 5$ which means that 80% of data for training and 20% for testing. Figs. 8 and 9 show the learned fuzzy sets for fuzzy variables *SA*, *LCD*, *SCL*, *STS*, and *SLP*, by applying GA (crossover rate / mutation rate = 0.9 / 0.1) and PSO with 84 particles learning mechanisms to learn 1000 and 3000 generations, respectively. Fig. 10 shows that learning 3000 generations performs better than the others for both GA and PSO. Additionally, the proposed PSO learning method has a better performance than GA learning.

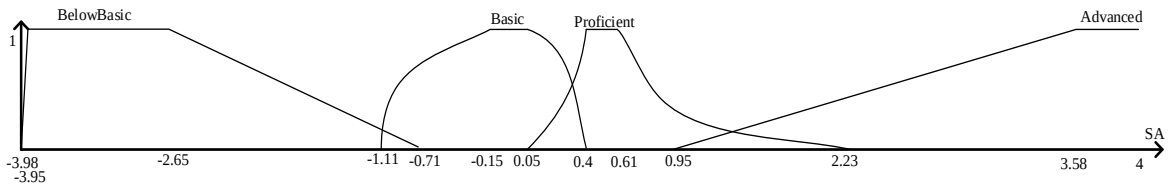

(a)

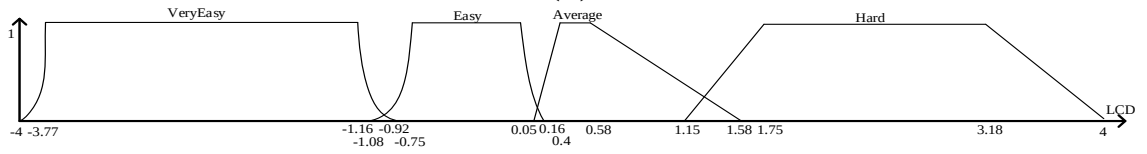

(b)

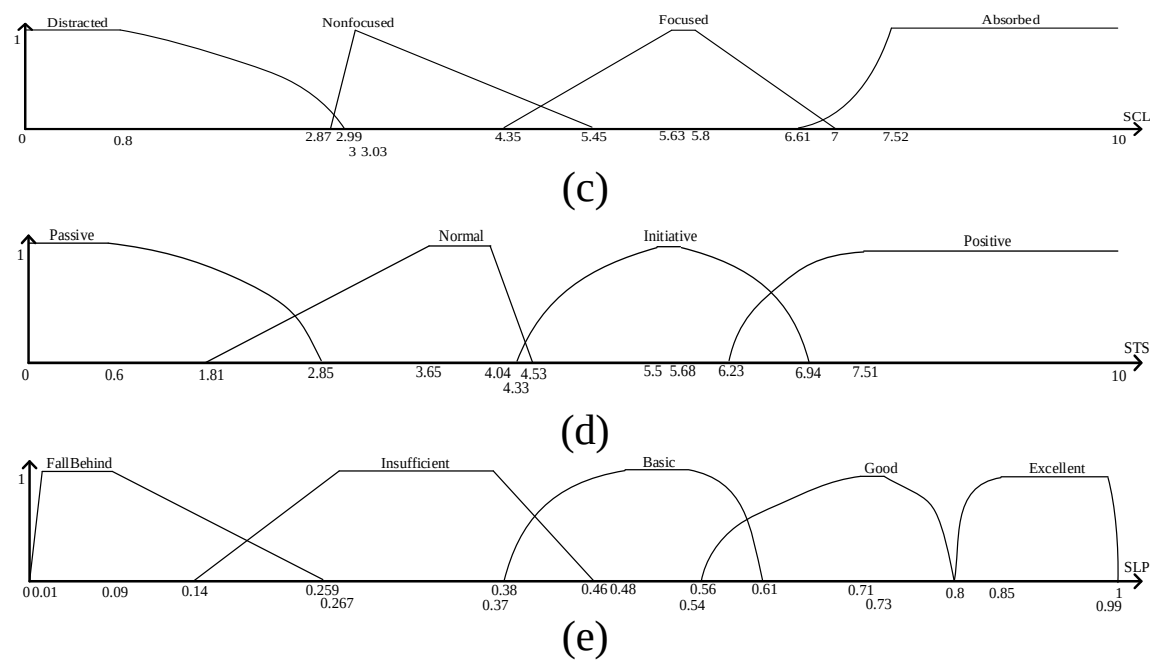

Fig. 8. FML variables after GA learning (a) *SA*, (b) *LCD*, (c) *SCL*, (d) *STS*, and (e) *SLP*.

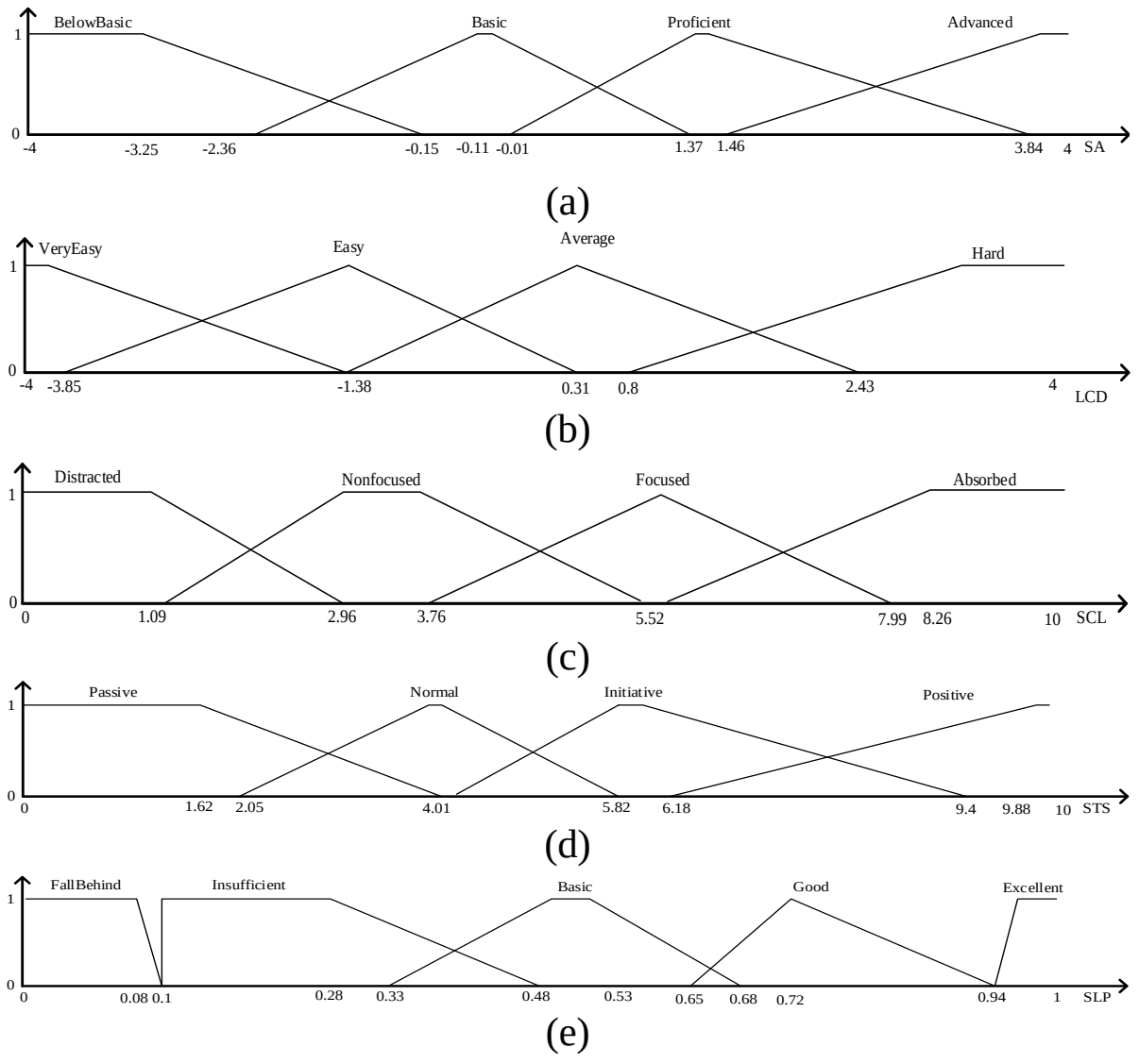

Fig. 9. FML variables after PSO learning (a) *SA*, (b) *LCD*, (c) *SCL*, (d) *STS*, and (e) *SLP*.

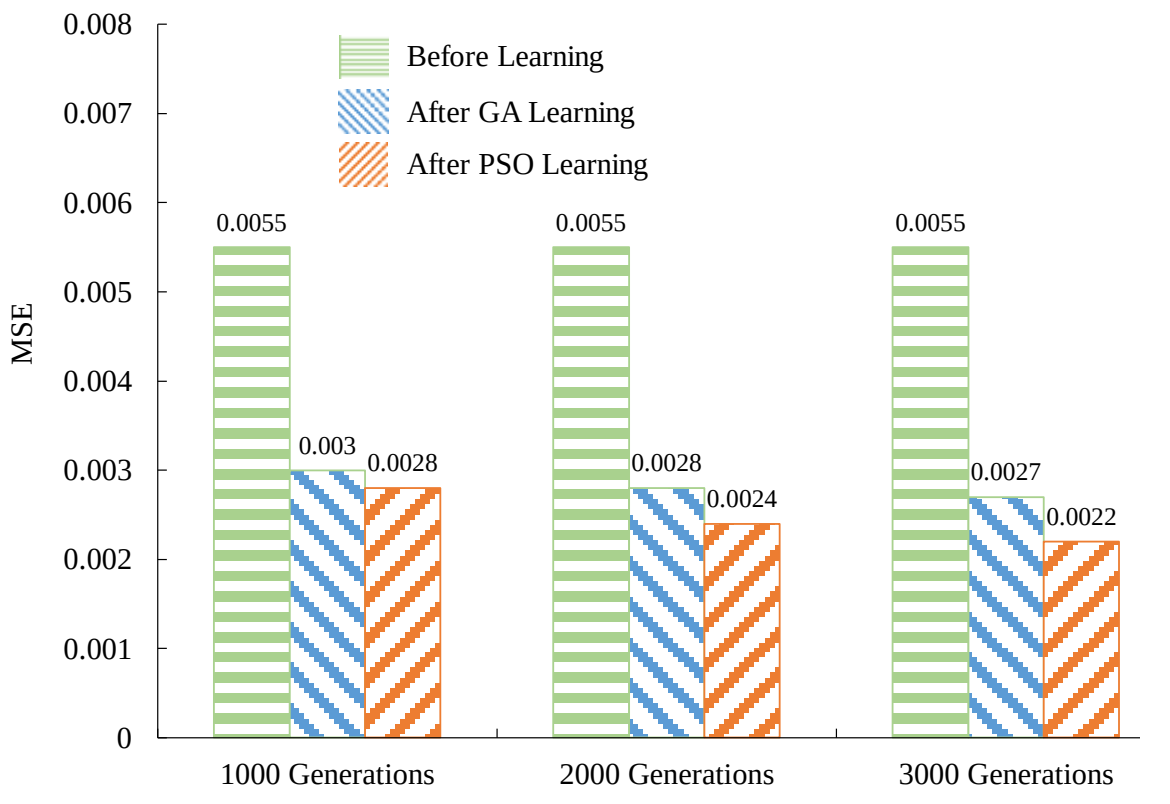

Fig. 10. MSE values of before learning, after GA learning, and after PSO learning with evolving 1000, 2000, and 3000 generations.

## B. *Part 2: Learning Content Recommendation*

The purpose of Part 2 of the experiments is to recommend learning contents for next students' learning by feeding the learned knowledge from Part 1 into the robot. We categorize the learning contents into four levels, including elementary, intermediate, high intermediate, and advanced levels. The input fuzzy variables are *Student Ability* (*SA*) and *Student Learning Performance* (*SLP*) and the output fuzzy variable is *Recommended Learning Content Rank* (*RLCR*) with 8 linguistic terms including last-grade high intermediate level (*LGHIL*), last-grade advanced level (*LGAL*), current-grade elementary level (*CGEL*), current-grade intermediate level (*CGIL*), current-grade high intermediate level (*CGHIL*), current-grade advanced level (*CGAL*), next-grade elementary level (*NGEL*), and next-grade intermediate level (*NGIL*). The range of *RLCR* is between -4 and +4 and it is the same as student's ability [20].

Table III shows partial knowledge base and rule base of learning content recommendation which is constructed according to the learned knowledge of PSO learning mechanism. The total number of fuzzy rules is 20 listed in Table IV. Table V lists partial input data that recommend the learning content rank and column $RLCR_{DO}$ is the desired output (DO). Columns $RLCR_{BLKB}$ and $RLCR_{ALKB}$ show partial inferred results when we extracted the parameters of input fuzzy variables *SA* and *SLP* from the before-learning and after-learning knowledge of PSO learning mechanism, respectively. Table V indicates that the robot agent with the learned knowledge recommends more suitable learning contents owing to an increase in accuracy from 78.75% to 87%.

TABLE III. KNOWLEDGE BASE OF PART-2 EXPERIMENT.

```xml
<knowledgeBase networkAddress="127.0.0.1">
    <fuzzyVariable name="SA" domainleft="-4" domainright="4" type="Input" accumulation="MAX" defuzzifier="COG" defaultValue="0.0"
    networkAddress="127.0.0.1">
        <fuzzyTerm name="BelowBasic" complement="false">
            <trapezoidShape param1="-4" param2="-4" param3="-3.25673655687434" param4="-0.152055386914671"/>
        </fuzzyTerm>
        <fuzzyTerm name="Basic" complement="false">
            <trapezoidShape param1="-2.36118982689676" param2="-0.11407888046308" param3="-0.0312655005335272" param4="1.37469875394825"/>
        </fuzzyTerm>
        <fuzzyTerm name="Proficient" complement="false">
            <trapezoidShape param1="-0.0103482154262207" param2="1.39281163804229" param3="1.46280282083579" param4="3.84407401339944"/>
        </fuzzyTerm>
        <fuzzyTerm name="Advanced" complement="false">
            <trapezoidShape param1="1.46883018373859" param2="3.86873126906892" param3="4" param4="4"/>
        </fuzzyTerm>
    </fuzzyVariable>
    <fuzzyVariable name="SLP" domainleft="0.0" domainright="1" type="Input" accumulation="MAX" defuzzifier="COG" defaultValue="0.0"
    networkAddress="127.0.0.1">
        <fuzzyTerm name="FallBehind" complement="false">
            <trapezoidShape param1="0" param2="0" param3="0.0897977479898309" param4="0.105200339370544"/>
        </fuzzyTerm>
        <fuzzyTerm name="Insufficient" complement="false">
            <trapezoidShape param1="0.105190543878888" param2="0.106059771230976" param3="0.284702332026263" param4="0.481276171863193"/>
        </fuzzyTerm>
        <fuzzyTerm name="Basic" complement="false">
            <trapezoidShape param1="0.332579982777035" param2="0.500997862520704" param3="0.538777582676206" param4="0.680045820686313"/>
        </fuzzyTerm>
        <fuzzyTerm name="Good" complement="false">
            <trapezoidShape param1="0.650501527786677" param2="0.725255936775388" param3="0.728011748758261" param4="0.944273821511864"/>
        </fuzzyTerm>
        <fuzzyTerm name="Excellent" complement="false">
            <trapezoidShape param1="0.944087727920405" param2="0.952713781791984" param3="1" param4="1"/>
        </fuzzyTerm>
    </fuzzyVariable>
    <fuzzyVariable name="RLCR" domainleft="-4" domainright="4" type="Output" accumulation="MAX" defuzzifier="COG" defaultValue="0.0"
    networkAddress="127.0.0.1">
        <fuzzyTerm name="LGHIL" complement="false">
            <trapezoidShape param1="-4" param2="-4" param3="-3.5" param4="-3"/>
        </fuzzyTerm>
        <fuzzyTerm name="LGAL" complement="false">
            <trapezoidShape param1="-3.5" param2="-3" param3="-2.5" param4="-2"/>
        </fuzzyTerm>
        <fuzzyTerm name="CGEL" complement="false">
            <trapezoidShape param1="-2.5" param2="-2" param3="-1.5" param4="-1"/>
        </fuzzyTerm>
        <fuzzyTerm name="CGIL" complement="false">
            <trapezoidShape param1="-1.5" param2="-1" param3="-0.5" param4="0"/>
        </fuzzyTerm>
        <fuzzyTerm name="CGHIL" complement="false">
            <trapezoidShape param1="-0.5" param2="0" param3="0.5" param4="1"/>
        </fuzzyTerm>
        <fuzzyTerm name="CGAL" complement="false">
            <trapezoidShape param1="0.5" param2="1" param3="1.5" param4="2"/>
        </fuzzyTerm>
        <fuzzyTerm name="NGEL" complement="false">
            <trapezoidShape param1="1.5" param2="2" param3="2.5" param4="3"/>
        </fuzzyTerm>
        <fuzzyTerm name="NGIL" complement="false">
            <trapezoidShape param1="2.5" param2="3" param3="4" param4="4"/>
        </fuzzyTerm>
    </fuzzyVariable>
</knowledgeBase>
```

TABLE IV. FUZZY RULES OF RECOMMENDING LEARNING CONTENT.

| No. | *SA* | *SLP* | *RLCR* | No. | *SA* | *SLP* | *RLCR* |
|---|---|---|---|---|---|---|---|
| 1 | Below Basic | Fall Behind | LGHIL | 11 | Proficient | Fall Behind | CGIL |
| 2 | Below Basic | Insufficient | LGAL | 12 | Proficient | Insufficient | CGHIL |
| 3 | Below Basic | Basic | LGAL | 13 | Proficient | Basic | CGAL |
| 4 | Below Basic | Good | CGEL | 14 | Proficient | Good | CGAL |
| 5 | Below Basic | Excellent | CGIL | 15 | Proficient | Excellent | NGEL |
| 6 | Basic | Fall Behind | LGAL | 16 | Advanced | Fall Behind | CGAL |
| 7 | Basic | Insufficient | CGEL | 17 | Advanced | Insufficient | CGAL |
| 8 | Basic | Basic | CGIL | 18 | Advanced | Basic | NGEL |
| 9 | Basic | Good | CGHIL | 19 | Advanced | Good | NGIL |
| 10 | Basic | Excellent | CGAL | 20 | Advanced | Excellent | NGIL |

TABLE V. PARTIAL INPUT DATA.

| No | *SA* | *SLP* | $RLCR_{DO}$ | $RLCL_{BLKB}$ | $RLCR_{ALKB}$ |
|---|---|---|---|---|---|
| 1 | -1.43 | 0.111 | -1.99067 | -3.611 | -2.248 |
| 2 | -1.03 | 0.167 | -1.57467 | -2.997 | -2.073 |
| 3 | -2.23 | 0.098 | -2.55867 | -2.985 | -2.67 |
| 4 | -1.88 | 0.11 | -2.29333 | -3.611 | -2.472 |
| 5 | -3.74 | 0.113 | -3.52533 | -3.611 | -3.611 |
| 6 | -2.87 | 0.116 | -2.93733 | -3.611 | -3.599 |
| 7 | -1.68 | 0.153 | -2.04533 | -2.791 | -2.369 |
| 8 | -0.97 | 0.117 | -1.668 | -3.024 | -2.049 |
| 9 | -1.5 | 0.105 | -2.05333 | -3.48 | -3.072 |
| 10 | -2.65 | 0.112 | -2.80133 | -3.611 | -2.75 |
| | | | ⋮ | | |
| 396 | 2.87 | 0.903 | 2.988 | 3.314 | 2.439 |
| 397 | 3.71 | 0.902 | 3.545333 | 3.367 | 2.735 |
| 398 | 1.43 | 0.803 | 1.761333 | 3.079 | 1.25 |
| 399 | 1.61 | 0.85 | 2.006667 | 3.367 | 1.441 |
| 400 | 1.57 | 0.907 | 2.132 | 3.339 | 1.5533 |
| Accuracy | | | | 78.75% | 87% |

### C. Part 3: FML-based Intelligent Agent for Student Learning

In Part 3 of the experiments, we deployed two different robots, Palro and Zenbo, with the proposed FML-based intelligent agent to an elementary school for students and robot co-learning. The involved fourth-grade elementary students were divided into two groups. Each group contains four students with different levels of ability. The purpose of group learning is to hope that students can learn together with the robots by teamwork and that students with stronger learning ability can guide weaker students along with the robot's assistance. Fig 11 shows the grouping learning diagram and actual teaching situation that the involved four-grade students and robot co-learning mathematics in the class in Kaohsiung, Taiwan on Nov. 25 and Dec. 23, 2017. Group A learns mathematics together with Palro and Zenbo but Group B does with Palro and iPads.

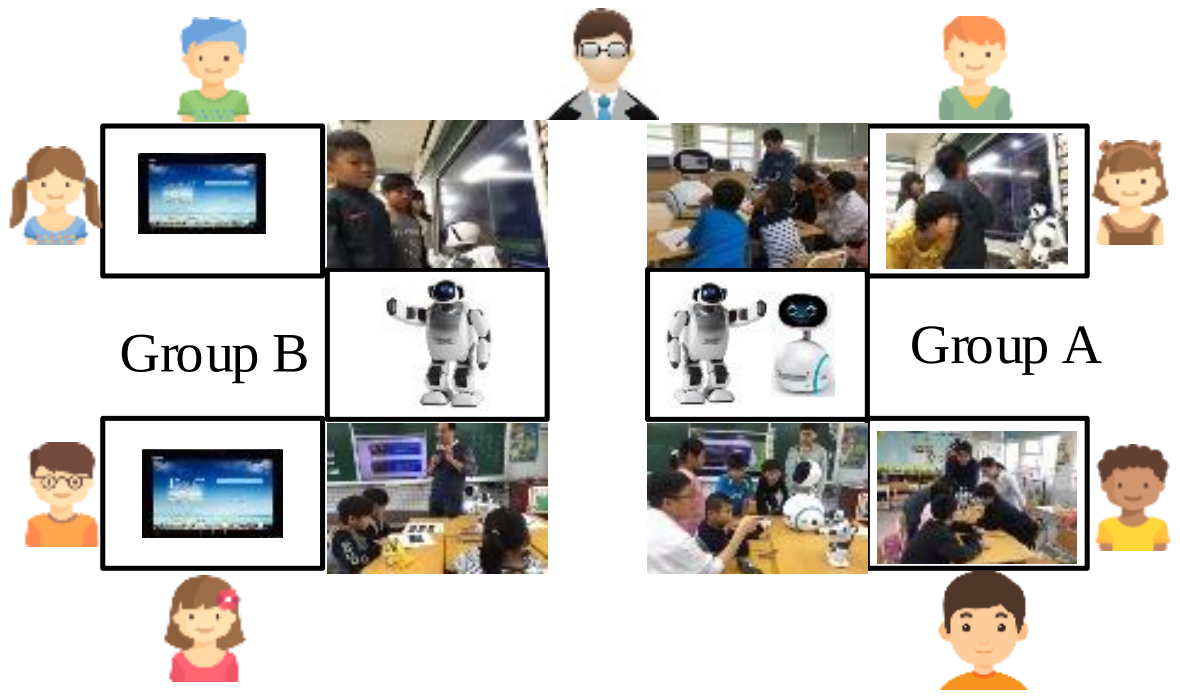

Fig. 11. Grouping learning diagram and actual teaching situation in the class.

First deploy is to learn the concepts of the number line on Nov. 25, 2017 through play. The total number of the items is 22. The bigger the item number, the harder the item. The students input their response data to shoot the target of the number line shown on the screen of Zenbo (Group A) or iPad (Group B). Meanwhile, Palro provided students with some hints when they failed to hit the target but cheered for them when they made it. In addition, Palro will provide different levels of hints according to the number of incorrect answer. The more number of incorrect answer, the more detailed the hints. Moreover, the students cannot do next item until they correctly answer the current item. Fig. 12 shows the average distance between two points (students' response data and correct answer), on a number line from Item 1 to Item 22. If the average distance is small, then the students' learning performance is good. We observe that Group B co-learns better with the robots than Group A.

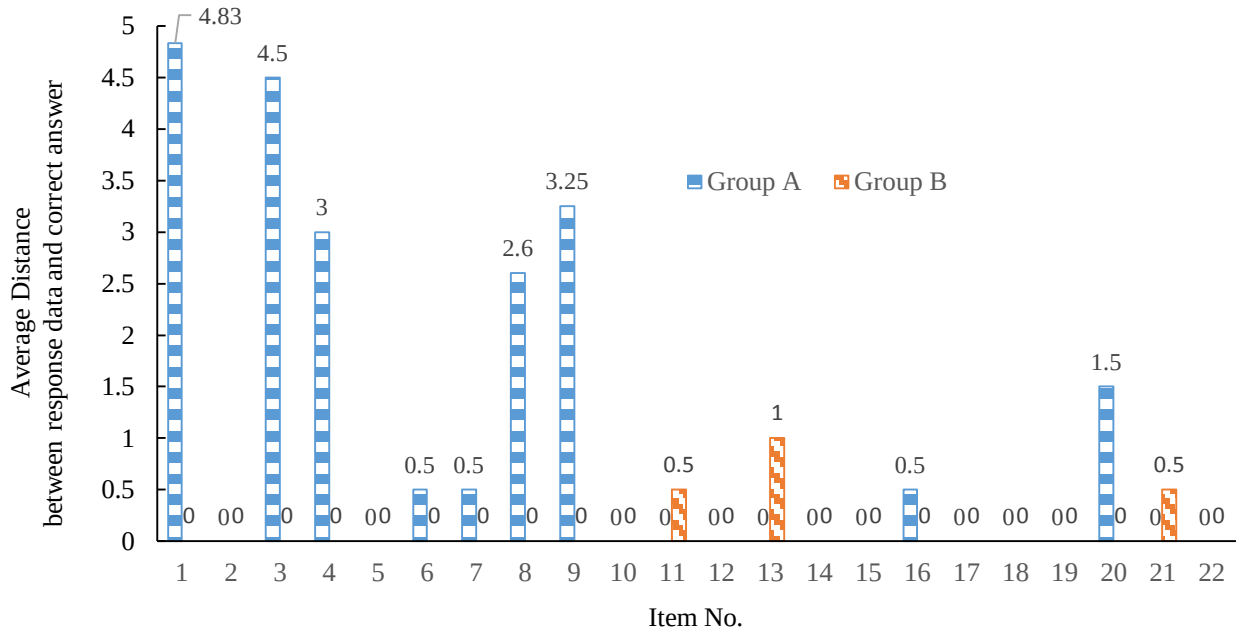

Fig. 12. Average distance between two points (students' response data and correct answer) on a number line for Items 1−22 for the study on Nov. 25, 2017.

The second study was about learning the concept of groups of numbers on Dec. 23, 2017 through play. Palro and Zenbo also play the similar role to the first study on Nov. 25, 2017. We used three monsters, including MonsterA, MonsterB, and MonsterC, to represent the scores 8, 12, and −4, respectively. The involved students hit the exact number of monsters to complete their mission. For example, if their mission is to get score 56 by hitting three kinds of monsters, they can hit three MonsterAs, one MonsterB, and one MonsterC to get score $3\times8 + 3\times12 - 1 \times 4 = 56$. Their challenged difficulty is divided into three levels, including *Easy*, *Average*, and *Hard*. Each challenge has different numbers of missions. However, there is an upper bound of the number of making a response to each mission. Table VI shows the obtained score for each level. The score is calculated according to how many times students make a response to the mission and whether they successfully complete the mission or not in the end. The total score is bounded in [0, 29].

TABLE VI. SCORE BASED ON HOW MANY TIMES STUDENTS TRY TO MAKE A RESPONSE.

| Times | Challenge Level | | |
|---|---|---|---|
| | Easy | Average | Hard |
| 1 | 2 | 3 | 3 |
| 2 | 1 | 2 | 2 |
| 3 | 0 | 1 | 1 |
| 4 | 0 | 0 | 0 |

Fig. 13 shows the number of making a response of each mission for Groups A and B on Dec. 23, 2017. The first mission of each challenge is to test if students fully understand their mission for the current challenge. Observe Fig. 13 that understanding their mission of the *Average* and *Hard* challenges is the most difficult one for both Groups A and B when we compare to the other missions. The symbol "×" of Fig. 13 denotes that the involved students failed this mission in the end. The total acquired score of Groups A and B is 23 and 22, respectively, which means that two-group students perform equally and grouping learning with the robot is feasible for the involved students.

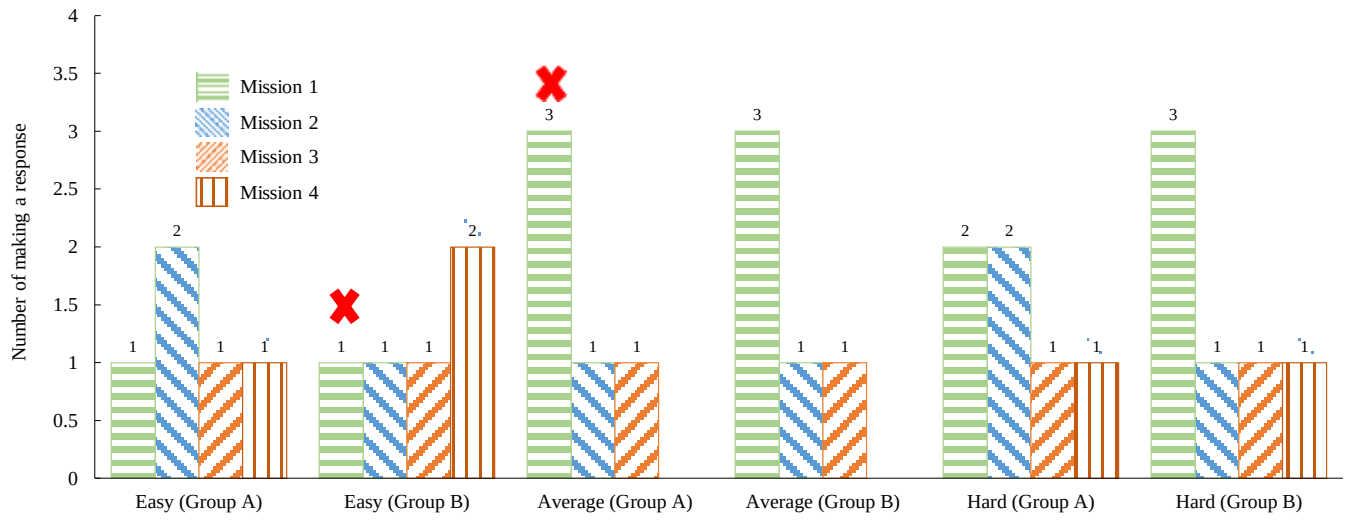

Fig. 13. Number of making a response of each mission for Groups A and B on Dec. 23, 2017.

## V. Conculsion and Future Work

This paper presents an FML-based intelligent agent for students and robot co-learning. The student learning performance ontology and the student with robot co-learning ontology are proposed for the intelligent agent. In addition, the machine learning mechanism, including GA and PSO, are also adopted for the knowledge base refinement. The machine-human co-learning model is established to help various students learn the mathematical concepts based on their learning ability and performance. Experimental results show that learning with the robot is helpful for the involved students. In the future, the intelligent agent with different robots will be deployed in various learning environments to help more students' learning.

## Acknowledgment

The authors would like to thank Han-Ru Liu, Jou-Te Tsai, master students of Dept. of Education, National University of Tainan (NUTN), faculty of San Pi Elementary School in Kaohsiung for their help with introducing the developed learning contents into the teaching field. Finally, we also would like to thank the involved students of San Pi Elementary School.